\pdfoutput=1

\documentclass[runningheads,a4paper]{llncs}

\usepackage{amssymb}
\usepackage{graphicx}
\usepackage{epstopdf}
\usepackage{epsfig}
\usepackage{graphicx}
\usepackage{amsmath}
\usepackage{amsfonts}
\usepackage{array,booktabs}
\usepackage{nameref}
\usepackage{caption}
\usepackage{cite}
\usepackage[breaklinks=true,bookmarks=false]{hyperref}
\usepackage[linesnumbered,ruled,vlined]{algorithm2e}
\usepackage{multirow}

\usepackage{url}
\urldef{\mailsa}\path|{...,|
\urldef{\mailsb}\path|...,|
\urldef{\mailsc}\path|...}@springer.com|    
\newcommand{\keywords}[1]{\par\addvspace\baselineskip
\noindent\keywordname\enspace\ignorespaces#1}

\SetAlFnt{\scriptsize}
\begin{document}

\mainmatter  % start of an individual contribution

\title{DOTE: Dual cOnvolutional filTer lEarning for Super-Resolution and Cross-Modality Synthesis in MRI}

\titlerunning{DOTE: Dual cOnvolutional filTer lEarning}

\author{Yawen Huang$^1$,
	Ling Shao$^2$,
	Alejandro F. Frangi$^1$}
%index{Huang, Yawen 
%	   Shao, Ling 
%	   F. Frangi, Alejandro}

%\thanks{Please note that the LNCS Editorial assumes that all authors have used
%the western naming convention, with given names preceding surnames. This determines
%the structure of the names in the running heads and the author index.}%
%\and Ursula Barth\and Ingrid Haas\and Frank Holzwarth\and\\
%Anna Kramer\and Leonie Kunz\and Christine Rei\ss\and\\
%Nicole Sator\and Erika Siebert-Cole\and Peter Stra\ss er}
%

\authorrunning{Y. Huang et al.}

\institute{$^1$Department of Electronic and Electrical Engineering, The University of Sheffield, UK\\
	$^2$School of Computing Sciences, University of East Anglia, UK\\
	{\tt\small \{yhuang36, a.frangi\}@sheffield.ac.uk, ling.shao@uea.ac.uk}}


\toctitle{DOTE: Dual cOnvolutional filTer lEarning for Super-Resolution and Cross-Modality Synthesis in MRI}
\tocauthor{Y. Huang et al.}
\maketitle

\begin{abstract}
Cross-modal image synthesis is a topical problem in medical image computing. Existing methods for image synthesis are either tailored to a specific application, require large scale training sets, or are based on partitioning images into overlapping patches. In this paper, we propose a novel Dual cOnvolutional filTer lEarning (DOTE) approach to overcome the drawbacks of these approaches. We construct a closed loop joint filter learning strategy that generates informative feedback for model self-optimization. Our method can leverage data more efficiently thus reducing the size of the required training set. We extensively evaluate DOTE in two challenging tasks: image super-resolution and cross-modality synthesis. The experimental results demonstrate superior performance of our method over other state-of-the-art methods.
\keywords{Dual learning $\cdot$ convolutional sparse coding $\cdot$ 3D $\cdot$ multi-modal $\cdot$ image synthesis $\cdot$ MRI.}
\end{abstract}

\section{Introduction}
\label{intro}
In medical image analysis, it is sometimes convenient or necessary to infer an image from one modality or resolution from another image modality or resolution for better disease visualization, prediction and detection purposes. A major challenge of cross-modality image segmentation or registration comes from the differences in tissue appearance or spatial resolution in images arising from different physical acquisition principles or parameters, which translates into the difficulty to represent and relate these images. Some existing methods tackle this problem by learning from a large amount of registered images and constraining pairwise solutions in a common space. In general, one would desire to have high-resolution (HR) three-dimensional Magnetic Resonance Imaging (MRI) with near isotropic voxel resolution as opposed to the more common image stacks of multiple 2D slices for accurate quantitative image analysis and diagnosis. Multi-modality imaging can generate tissue contrast arising from various anatomical or functional features that present complementary information about the underlying organ. Acquiring low-resolution (LR) single-modality images, however, is not uncommon. 

To solve the above problems, super-resolution (SR) \cite{Yang,Timofte} reconstruction is carried out for recovering an HR image from its LR counterpart, and cross-modality synthesis (CMS) \cite{Vemulapalli} is proposed for synthesizing target modality data from available source modality images. Generally, these methods have explored image priors from either internal similarities of image itself \cite{Rousseau} or external data support \cite{Zeyde}, to construct the relationship between two modalities. Although these methods achieve remarkable results, most of them suffer from the fundamental limitations associated with large scale pairwise training sets or patch-based overlapping mechanism. Specifically, a large amount of multi-modal images is often required to learn a sufficiently expressive dictionaries/networks. However, this is impractical since collecting medical images is very costly and limited by many factors. On the other side, patch-based methods are subjected to inconsistencies introduced during the fusion process that takes place in areas where patches overlap.

To deal with the bottlenecks of training data and patch-based implementation, we develop a dual convolutional filter learning (DOTE) method with an application to neuroimaging that investigates data (in both source and target modalities from the same set of subjects) in a more effective way, and solves image SR and CMS problems respectively. The contributions of this work are mainly in four aspects: (1) We present a unified model (DOTE) for any cross-modality image synthesis problem; (2) The proposed method can efficiently reduce the amount of training data needed from the model, by generating abundant feedbacks from dual mapping functions during the training process; (3) Our method integrates feature learning and mapping relation in a closed loop for self-optimization. Local neighbors are preserved intrinsically by directly working on the whole images; (4) We evaluate DOTE on two datasets in comparison with stat-of-the-art methods. Experimental results demonstrate superior performance of DOTE over these approaches.

\begin{figure}[b]
	\vspace{-0.2cm}
	\setlength{\abovecaptionskip}{-0.01cm}
	\centering
	\includegraphics[width=0.9\linewidth]{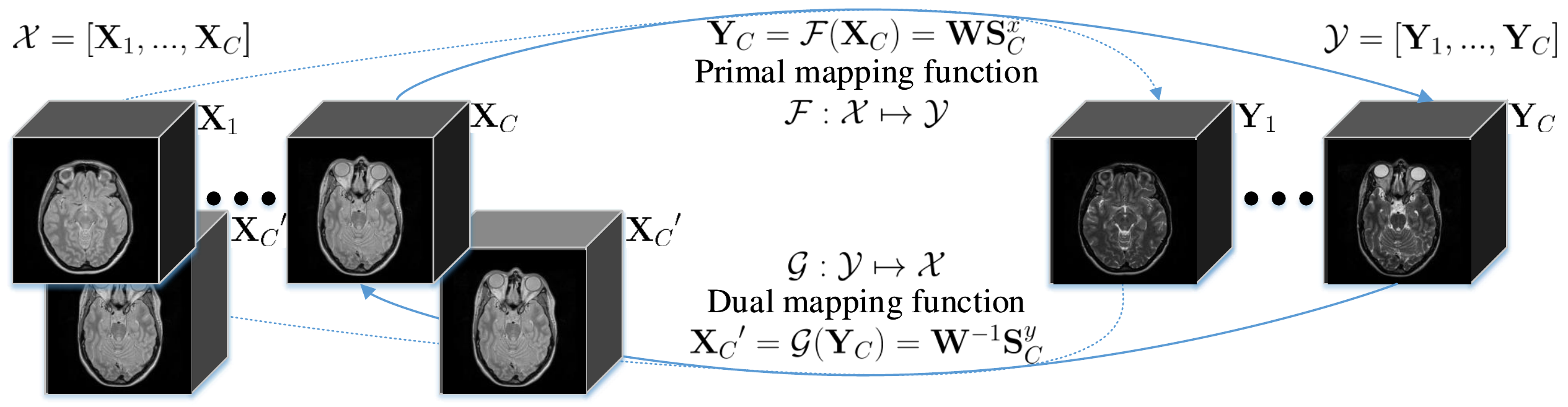}
	\caption{Flowchart of the proposed method for MRI cross-modality synthesis.}
	\label{fig1}
\end{figure}
\section{Method}
\subsection{Background}
\label{background}
\textbf{Convolutional Sparse Coding} (CSC) remedies a fundamental drawback of conventional patch-based sparse representation methods by modeling shift invariance for consistent approximation of local neighbors on whole images. Instead of decomposing the vector as the multiplication of dictionary atoms and the coded coefficients, CSC provides a more elegant way to model local interactions. That is, by representing an image as the summation of convolutions of the sparsely distributed feature maps and the corresponding filters. Concretely, given an $m \times n$ image $\mathbf{x}$ in vector form, the problem of learning a set of vectorized filters　for sparse feature maps is solved by minimizing the objective function that combines the convolutional least-squares term and the $l_1$-norm penalty on the representations:
\begin{equation}
\begin{aligned}
\label{eq１}
\arg \min_{\mathbf{f},\mathbf{s}} \frac{1}{2}\left \| \mathbf{x}-\sum_{k=1}^{K}\mathbf{f}_{k}\ast \mathbf{s}_{k} \right \|_{2}^{2}+\lambda \sum_{k=1}^{K}\left \| \mathbf{s}_{k} \right \|_{1}\\
s.t. \; \left \| \mathbf{f}_{k} \right \|_{2}^{2}\leq 1 \;\; \forall k=\left \{ 1,...,K \right \},
\end{aligned}
\end{equation}

\noindent where $\mathbf{f}_{k} \in \mathbf{f}=\left [ \mathbf{f}_{1}^{T},...,\mathbf{f}_{K}^{T} \right ]^{T}$ is the $k$-th $d \times d$ filter, $\ast$ denotes the 2D convolution operator, $\mathbf{z}_{k} \in \mathbf{z}=\left [ \mathbf{z}_{1}^{T},...,\mathbf{z}_{K}^{T} \right ]^{T}$ refers to the sparse feature map corresponding to $\mathbf{f}_{k}$ with size $\left ( m+d-1 \right )\times \left ( n+d-1 \right )$ to approximate $\mathbf{x}$, and $\lambda$ is a regularization parameter. The problem in Eq. (\ref{eq１}) can be efficiently and explicitly solved in the Fourier domain, derived within an Alternating Direction Method of Multipliers (ADMM) framework \cite{Bristow}.

\textbf{Dual Learning} (DL) \cite{He} is a new learning paradigm that translates the input model by forming a closed loop between source and target domains to generate informative feedbacks. Specifically, for any dual tasks (e.g., $A \leftrightarrow B$) DL strategy appoints $A \rightarrow B$ as the primary task and the other $A \leftarrow B$ as the dual task, and forces them learning from each other to produce the pseudo-input ${A}'$. It can achieve the comparable performance through iteratively updating and minimizing the reconstruction error $A-{A}'$ that helps maximize the use of data. Therefore, making the learning-based methods have less dependent on the large number of training data.

\textbf{Problem Formulation}: The cross-modality image synthesis problem can be formulated as: given an 3D image $\mathbf{X}$ of modality $\mathcal{M}_{1}$, the task is to infer from $\mathbf{X}$ a target 3D image $\mathbf{Y}$ that approximates to the ground truth of modality $\mathcal{M}_{2}$. Let $\mathcal{X}=\left [ \mathbf{X}_{1}, ...,\mathbf{X}_{C}\right ] \in \mathbb{R}^{m\times n\times z\times C}$ be a set of images of modality $\mathcal{M}_{1}$ in the source domain, and $\mathcal{Y}=\left [ \mathbf{Y}_{2}, ...,\mathbf{Y}_{C}\right ] \in \mathbb{R}^{m\times n\times z\times C}$ be a set of images of modality $\mathcal{M}_{2}$ in the target domain. $m$, $n$ are the dimensions of axial view of the image, and $z$ denotes the size of image along the z-axis, while $C$ is the numbers of elements in the training sets. Each pair of $\left \{ \mathbf{X}_{i},\mathbf{Y}_{i} \right \}$ $\forall i=\left \{ 1,...,C \right \}$ are registered. To bridge image appearances across different modalities while preserving the intrinsic local interactions (i.e., intra-domain consistency), we propose a method based on CSC to jointly learn a pair of filters $\mathbf{F}^{x}$ and $\mathbf{F}^{Y}$. Moreover, inspired by the DL strategy, we form a closed loop between both domains and assume that there exists a primal mapping function $\mathcal{F}\left ( \cdot \right )$ from $\mathcal{X}$ to $\mathcal{Y}$ for relating and predicting from one another. We also assume there exists a dual mapping function $\mathcal{G}\left ( \cdot \right )$ from $\mathcal{Y}$ to $\mathcal{X}$ to generate feedbacks for model self-optimization. Experimentally, we investigate human brain MRI and apply our method to two cross-modality synthesis tasks, i.e., image SR and CMS. An overview of our method is depicted in Fig. \ref{fig1}. \textbf{Notation}: Matrices and 3D images are written in bold uppercase (e.g., image $\mathbf{X}$), vectors and vectorized 2D images in bold lowercase (e.g.,  filter $\mathbf{f}$) and scalars in lowercase (e.g., element $k$).

\subsection{Dual Convolutional Filter Learning}
Inspired by CSC (cf. Sec. \ref{background}) and the benefits of conventional coupled sparsity, we propose a dual convolutional filter learning (DOTE) model, which extends the original CSC formulation into a DL strategy and joint representation into a unified framework. More specifically, given $\mathcal{X}$ together with the corresponding $\mathcal{Y}$ for training, in order to facilitate a joint mapping, we associate the sparse feature maps of each registered data pair $\left \{ \mathbf{X}_{i},\mathbf{Y}_{i} \right \}_{i=1}^{C}$ by constructing a forward mapping function $\mathcal{F}: \mathcal{X} \mapsto \mathcal{Y}$ with $\mathbf{Y}=\mathcal{F}\left ( \mathbf{X} \right )$. Since such cross-modality synthesis problem satisfies a dual-learning mechanism, we further leverage the duality of the bidirectional transformation between the two domains. That is, by establishing a dual mapping function $\mathcal{G}: \mathcal{Y} \mapsto \mathcal{X}$ with $\mathbf{Y}=\mathcal{G}\left ( \mathbf{X} \right )$. Incorporating feature maps representing and the above closed-loop mapping functions, we can thus derive the following objective function:
\begin{equation}
\begin{aligned}
\label{obj}
\arg \min_{\mathbf{F}^{x},\mathbf{F}^{y},\mathbf{S}^{x},\mathbf{S}^{y}, \mathbf{W}} \frac{1}{2} \left \| \mathbf{X}-\sum_{k=1}^{K}\mathbf{F}_{k}^{x}\ast \mathbf{S}_{k}^{x} \right \|_{F}^{2}+\frac{1}{2} \left \| \mathbf{Y}-\sum_{k=1}^{K}\mathbf{F}_{k}^{y}\ast \mathbf{S}_{k}^{y} \right \|_{F}^{2}+\gamma \sum_{k=1}^{K}\left \| \mathbf{W}_{k} \right \|_{F}^{2}\\
+\lambda\left ( \sum_{k=1}^{K}\left \| \mathbf{S}_{k}^{x} \right \|_{1} +\sum_{k=1}^{K}\left \| \mathbf{S}_{k}^{y} \right \|_{1} \right )+\beta \left ( \sum_{k=1}^{K}\left \| \mathbf{S}_{k}^{y}-\mathbf{W}_{k}\mathbf{S}_{k}^{x} \right \|_{F}^{2} +\sum_{k=1}^{K}\left \| \mathbf{S}_{k}^{x}-\mathbf{W}_{k}^{-1}\mathbf{S}_{k}^{y} \right \|_{F}^{2} \right)\\
s.t. \; \left \| \mathbf{f}_{k}^{x} \right \|_{2}^{2}\leq 1 \; \left \| \mathbf{f}_{k}^{y} \right \|_{2}^{2}\leq 1 \;\; \forall k=\left \{ 1,...,K \right \}.
\end{aligned}
\end{equation}

\noindent where $\mathbf{S}_{k}^{x}$ and $\mathbf{S}_{k}^{y}$ take the role of the $k$-th sparse feature maps that approximate data $\mathbf{X}$ and $\mathbf{Y}$ when convolved with the $k$-th filters $\mathbf{F}_{k}^{x}$ and $\mathbf{F}_{k}^{y}$ of a fixed spatial support, $k=1,...,K$. $\left \| \cdot  \right \|_{F}$ is a Frobenius norm chosen to induce the convolutional least squares approximation, and $\ast$ is represented as a 3D convolution operator, while $\lambda$, $\beta$, $\gamma$ are the regularization parameters. Particularly, dual mapping functions $\mathcal{F}\left ( \mathbf{S}_{k}^{x}, \mathbf{W}_{k}\right )=\mathbf{W}_{k} \mathbf{S}_{k}^{x}$ and $\mathcal{G}\left ( \mathbf{S}_{k}^{y}, \mathbf{W}_{k}^{-1}\right )=\mathbf{W}_{k}^{-1} \mathbf{S}_{k}^{y}$ are used to relate the sparse feature maps of $\mathbf{X}$ and $\mathbf{Y}$ over $\mathbf{F}^{x}$ and $\mathbf{F}^{y}$. They are done by solving two sets of least squares terms (i.e., $\sum_{k=1}^{K}(\left \| \mathbf{S}_{k}^{y}-\mathbf{W}_{k}\mathbf{S}_{k}^{x} \right \|_{F}^{2}\\+\sum_{k=1}^{K} \left \| \mathbf{S}_{k}^{x}-\mathbf{W}_{k}^{-1}\mathbf{S}_{k}^{y} \right \|_{F}^{2})$ with respect to the linear projections.

\subsection{Optimization}
Similar to classical dictionary learning methods, the objective function in Eq. (\ref{obj}) is not simultaneously convex with respect to the learned filter pairs,the sparse feature maps and the mapping. Instead, we divide the proposed method into three sub-problems: learning $\mathbf{S}^{x}$, $\mathbf{S}^{y}$, training $\mathbf{F}^{x}$, $\mathbf{F}^{y}$, and updating $\mathbf{W}$.

\textbf{Computing sparse feature maps}: We first initialize the filters $\mathbf{F}^{x}$, $\mathbf{F}^{y}$ as two random matrices and the mapping $\mathbf{W}$ as an identity matrix, then fix them for calculating the solutions of sparse feature maps $\mathbf{S}^{x}$, $\mathbf{S}^{y}$. As a result, the problem of Eq. (\ref{obj}) can be converted into two optimization sub-problems. Unfortunately, this cannot be solved under $l_{1}$ penalty without breaking rotation invariance. The resulting alternating algorithms \cite{Bristow} by introducing two auxiliary variables $\mathbf{U}$ and $\mathbf{V}$ enforce the constraint inherent in the splitting. In this paper, we follow \cite{Bristow} and solve the convolution subproblems in the Fourier domain within an ADMM optimization strategy:
\begin{equation}
\begin{aligned}
\label{opt1}
\min_{\mathbf{S}^{x}}\frac{1}{2} \left \| \hat{\mathbf{X}}-\sum_{k=1}^{K}\hat{\mathbf{F}}_{k}^{x}\odot  \hat{\mathbf{S}}_{k}^{x} \right \|_{F}^{2}+\lambda\sum_{k=1}^{K}\left \| \mathbf{U}_{k}^{x} \right \|_{1}+\beta\sum_{k=1}^{K}\left \| \hat{\mathbf{S}}_{k}^{y}-\mathbf{W}_{k}\hat{\mathbf{S}}_{k}^{x} \right \|_{F}^{2}\\
s.t.\; \left \| \mathbf{V}_{k}^{x} \right \|_{2}^{2}\leq 1, \; \mathbf{V}_{k}^{x}=\mathbf{V}\mathbf{\Phi}^{T}\hat{\mathbf{F}}_{k}^{x}, \; \mathbf{U}_{k}^{x}=\mathbf{S}_{k}^{x} \;\; \forall k=\left \{ 1,...,K \right \},\\
\min_{\mathbf{S}^{y}}\frac{1}{2} \left \| \hat{\mathbf{Y}}-\sum_{k=1}^{K}\hat{\mathbf{F}}_{k}^{y}\odot  \hat{\mathbf{S}}_{k}^{y} \right \|_{F}^{2}+\lambda\sum_{k=1}^{K}\left \| \mathbf{U}_{k}^{y} \right \|_{1}+\beta\sum_{k=1}^{K}\left \| \hat{\mathbf{S}}_{k}^{x}-\mathbf{W}_{k}^{-1}\hat{\mathbf{S}}_{k}^{y} \right \|_{F}^{2}\\
s.t.\; \left \| \mathbf{V}_{k}^{y} \right \|_{2}^{2}\leq 1, \; \mathbf{V}_{k}^{y}=\mathbf{V}\mathbf{\Phi}^{T}\hat{\mathbf{F}}_{k}^{y}, \; \mathbf{U}_{k}^{y}=\mathbf{S}_{k}^{y} \;\; \forall k=\left \{ 1,...,K \right \},
\end{aligned}
\end{equation}

\noindent where $\hat{}$ applied to any symbol denotes the frequency representations (i.e., Discrete Fourier Transform (DFT)). For instance, $\hat{\mathbf{X}} \leftarrow f(\mathbf{X})$ where $f(\cdot)$ is the Fourier transform operator. $\odot$ represents the component-wise product. $\mathbf{\Phi}^{T}$ is the inverse DFT matrix, and $\mathbf{V}$ projects a filter onto the small spatial support. The auxiliary variables $\mathbf{U}_{k}^{x}$, $\mathbf{U}_{k}^{y}$, $\mathbf{V}_{k}^{x}$ and $\mathbf{V}_{k}^{y}$ relax each of the CSC problems under dual mapping constraint by leading to several subproblem decompositions.

\textbf{Learning convolutional filters}: Like when solving for sparse feature maps, filter pairs can be learned similarly by setting $\mathbf{S}^{x}$, $\mathbf{S}^{y}$ and $\mathbf{W}$ fixed, and then learning $\mathbf{F}^{x}$ and $\mathbf{F}^{y}$ by minimizing
\begin{equation}
\begin{aligned}
\label{opt2}
\min_{\mathbf{F}^{x},\mathbf{F}^{y}} \frac{1}{2} \left \| \hat{\mathbf{X}}-\sum_{k=1}^{K}\hat{\mathbf{F}}_{k}^{x}\odot \hat{\mathbf{S}}_{k}^{x} \right \|_{F}^{2}+\frac{1}{2} \left \| \hat{\mathbf{Y}}-\sum_{k=1}^{K}\hat{\mathbf{F}}_{k}^{y}\odot \hat{\mathbf{S}}_{k}^{y} \right \|_{F}^{2}\\
s.t. \; \left \| \mathbf{f}_{k}^{x} \right \|_{2}^{2}\leq 1, \left \| \mathbf{f}_{k}^{y} \right \|_{2}^{2}\leq 1 \; \; \forall k=\left \{ 1,...,K \right \},
\end{aligned}
\end{equation}

Eq. (\ref{opt2}) can be solved by a one-by-one update strategy \cite{Wang} through an augmented Lagrangian method \cite{Bristow}.

\textbf{Updating mapping}: With fixed $\mathbf{F}^{x}$, $\mathbf{F}^{y}$, $\mathbf{S}^{x}$ and $\mathbf{S}^{y}$, we solve the following ridge regression problem for updating mapping $\mathbf{W}$:
\begin{equation}
\begin{aligned}
\label{opt3}
\min_{\mathbf{W}}\sum_{k=1}^{K}\left \| \mathbf{S}_{k}^{y}-\mathbf{W}_{k}\mathbf{S}_{k}^{x} \right \|_{F}^{2}+\left \| \mathbf{S}_{k}^{x}-\mathbf{W}_{k}^{-1}\mathbf{S}_{k}^{y} \right \|_{F}^{2}+\left (\frac{\gamma}{\beta}  \right )\sum_{k=1}^{K}\left \| \mathbf{W}_{k} \right \|_{F}^{2}.
\end{aligned}
\end{equation}

Particularly, the primal mapping function $\left \| \mathbf{S}_{k}^{y}-\mathbf{W}_{k}\mathbf{S}_{k}^{x} \right \|_{F}^{2}$ constructs an intrinsic mapping while the corresponding dual mapping function $\left \| \mathbf{S}_{k}^{x}-\mathbf{W}_{k}^{-1}\mathbf{S}_{k}^{y} \right \|_{F}^{2}$ is utilized to give feedbacks and further optimize the relationship between $\mathbf{S}_{k}^{x}$ and $\mathbf{S}_{k}^{y}$. Ideally (as the final solution), $\mathbf{S}_{k}^{y}=\mathbf{W}_{k}\mathbf{S}_{k}^{x}$, such that the problem in Eq. (\ref{opt3}) is reduced to $\min_{\mathbf{W}_k}\sum_{k=1}^{K}\left \| \mathbf{S}_{k}^{y}-\mathbf{W}_{k}\mathbf{S}_{k}^{x} \right \|_{F}^{2}+\left (\frac{\gamma}{\beta}  \right )\sum_{k=1}^{K}\left \| \mathbf{W}_{k} \right \|_{F}^{2}$ with the solution $\mathbf{W}=\mathbf{S}_{k}^{y}{\mathbf{S}_{k}^{x}}^{T}(\mathbf{S}_{k}^{x}{\mathbf{S}_{k}^{x}}^{T}+\frac{\gamma}{\beta} \mathbf{I})^{-1}$, where $\mathbf{I}$ is an identity matrix. We summarize the proposed DOTE method in the following Algorithm \ref{alg1}. 
\setlength{\textfloatsep}{2pt}
\begin{algorithm}[h]
	\SetAlFnt{\small\sf}
	\caption{DOTE algorithm} 
	\label{alg1}
	\SetAlgoLined
	\KwIn{Training data $\mathbf{X}$ and $\mathbf{Y}$, parameters $\lambda$, $\gamma$, $\sigma$.}
	Initialize $\mathbf{F}_{0}^{x}$, $\mathbf{F}_{0}^{y}$, $\mathbf{S}_{0}^{x}$, $\mathbf{S}_{0}^{y}$, $\mathbf{W}_{0}$, $\mathbf{U}_{0}^{x}$, $\mathbf{U}_{0}^{y}$, $\mathbf{V}_{0}^{x}$, $\mathbf{V}_{0}^{y}$.\\
	Perform FFT $\mathbf{S}_{0}^{x}\rightarrow \hat{\mathbf{S}}_{0}^{x}$, $\mathbf{S}_{0}^{y}\rightarrow \hat{\mathbf{S}}_{0}^{y}$, $\mathbf{F}_{0}^{x}\rightarrow \hat{\mathbf{F}}_{0}^{x}$, $\mathbf{F}_{0}^{y}\rightarrow \hat{\mathbf{F}}_{0}^{y}$, $\mathbf{U}_{0}^{x}\rightarrow \hat{\mathbf{U}}_{0}^{x}$, $\mathbf{U}_{0}^{y}\rightarrow \hat{\mathbf{U}}_{0}^{y}$, $\mathbf{V}_{0}^{x}\rightarrow \hat{\mathbf{V}}_{0}^{x}$, $\mathbf{V}_{0}^{y}\rightarrow \hat{\mathbf{V}}_{0}^{y}$.\\
	Let $\hat{\mathbf{S}}_{0}^{y} \leftarrow \mathbf{W}\hat{\mathbf{S}}_{0}^{x}$.\\
	\While{not converged}{
		Solve for $\hat{\mathbf{S}}_{k+1}^{x}$, $\hat{\mathbf{S}}_{k+1}^{y}$, $\hat{\mathbf{U}}_{k+1}^{x}$ and $\hat{\mathbf{U}}_{k+1}^{y}$ using (\ref{opt1}) with fixed filters and $\mathbf{W}_{k}$.\\
		Train $\hat{\mathbf{F}}_{k+1}^{x}$, $\hat{\mathbf{F}}_{k+1}^{y}$, $\hat{\mathbf{V}}_{k+1}^{x}$ and $\hat{\mathbf{V}}_{k+1}^{y}$ by (\ref{opt2}) with fixed feature maps and $\mathbf{W}_{k}$.\\
		Update $\mathbf{W}_{k+1}$ by (\ref{opt3}).\\
		Inverse FFT $\hat{\mathbf{F}}_{k+1}^{x}\rightarrow \mathbf{F}_{k+1}^{x}$, $\hat{\mathbf{F}}_{k+1}^{y}\rightarrow \mathbf{F}_{k+1}^{y}$.\\
	}
	\KwOut{$\mathbf{F}^{x}$, $\mathbf{F}^{y}$, $\mathbf{W}$.}
\end{algorithm}

\subsection{Synthesis}
Once the optimization is completed, we can obtain the learned filters $\mathbf{F}^{x}$, $\mathbf{F}^{y}$ and the mapping $\mathbf{W}$. We then apply the proposed model to synthesize images across different modalities (i.e., LR $\rightarrow$ HR and $\mathcal{M}_{1} \rightarrow \mathcal{M}_{2}$, respectively). Given a test image $\mathbf{X}^{t}$, we compute the sparse feature maps $\mathbf{S}^{tx}$ related to $\mathbf{F}^{x}$ by solving a single CSC problem like Eq. (\ref{eq１}): $\mathbf{S}^{tx}=\arg \min_{\mathbf{S}^{tx}} \frac{1}{2}\left \| \mathbf{X}^{t}-\sum_{k=1}^{K}\mathbf{F}_{k}^{x}\ast \mathbf{S}_{k}^{tx} \right \|_{2}^{2}+\lambda \sum_{k=1}^{K}\left \| \mathbf{S}_{k}^{tx} \right \|_{1}$. After that, we can synthesize the target modality image of $\mathbf{X}^{t}$ by the sum of $K$ target feature maps $\mathbf{S}_{k}^{ty}=\mathbf{W}\mathbf{S}_{k}^{tx}$ convolved with $\mathbf{F}_{k}^{y}$, i.e., $\mathbf{Y}^{t}=\sum_{k=1}^{K}\mathbf{F}_{k}^{y}\mathbf{S}_{k}^{ty}$.

\section{Experimental Results}
\textbf{Experimental Setup:} The proposed DOTE is validated on two datasets: IXI\footnote{http://brain-development.org/ixi-dataset/} (including 578 $256 \times 256 \times p$ MR healthy subjects) and NAMIC\footnote{http://hdl.handle.net/1926/1687} (involving 20 $128 \times 128 \times q$ subjects). In our experiments, we perform 4-fold cross-validation for testing. That is, selecting 144 subjects from IXI and 5 subjects from NAMIC, respectively, as our test data. Following~\cite{Yang,Wang}, the regularization parameters $\sigma$, $\lambda$, $\beta$, and $\gamma$ are empirically set to be 1, 0.05, 0.10, 0.15, respectively. The number of filters is set as 800 according to \cite{Gu}. Convergence towards primal feasible solution is proved in \cite{Bristow} by first converting Eq. (\ref{obj}) into two optimization sub-problems that involve two proxies $\mathbf{U}$, $\mathbf{V}$ and then solving them alternatively. DOTE converges after ca. 10 iterations. For the evaluation criteria, we adopt PSNR and SSIM indices to objectively assess the quality of our results. 

\begin{figure}[t!]
	\setlength{\abovecaptionskip}{-0.0cm}
	\setlength{\belowcaptionskip}{-0.25cm}
	\centering 
	\includegraphics[width=1\linewidth]{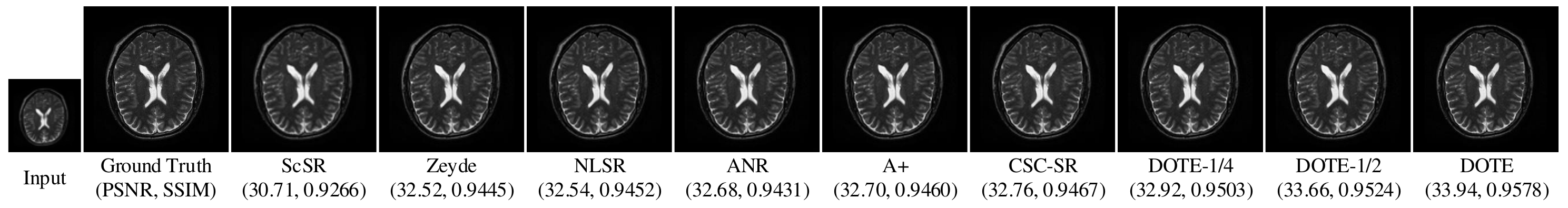} 
	\caption{Example SR results and the corresponding PSNRs and SSIMs.}
	\label{fig2} 
\end{figure}

\begin{figure}[t!]
	\setlength{\abovecaptionskip}{-0.0cm}
	\setlength{\belowcaptionskip}{-0.25cm}
	\centering 
	\includegraphics[width=1\linewidth]{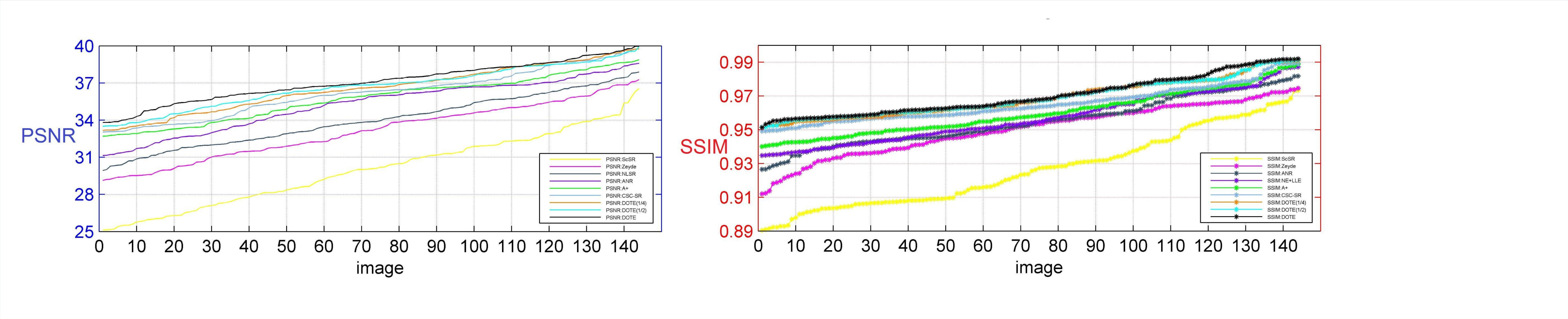} 
	\caption{Error measures of SR results on the IXI dataset.}
	\label{fig3} 
\end{figure}

\textbf{MRI Super-Resolution.} As we introduced in Section \ref{intro}, we first address image SR as one of cross-modality image synthesis. In this scenario, we investigate the T2-w images of the IXI dataset for evaluating and comparing DOTE with ScSR \cite{Yang}, A+ \cite{Timofte}, NLSR \cite{Rousseau}, Zeyde \cite{Zeyde}, ANR \cite{Timofte2}, and CSC-SR \cite{Gu}. Generally, LR images are generated by down-sampling  HR ground-truth images using bicubic interpolation. We perform image SR with scaling factor 2, and show visual results in Fig. \ref{fig2}. The quantitative results are reported in Fig. \ref{fig3}, while the average PSNRs and SSIMs for all 144 test subjects are shown in Table \ref{tab1}. The proposed model achieves the best PSNRs and SSIMs. Moreover, to validate our argument that DL-based self-optimization strategy is beneficial and requires less training data, we compare $\textrm{DOTE}_{\textrm{nodual}}$ (removing dual mapping term) and DOTE under different training data size (i.e., $\frac{1}{4},\frac{1}{2},\frac{3}{4}$ of the original dataset). The results are listed in Table \ref{tab2}. From Table \ref{tab2}, we see that DOTE is always better than $\textrm{DOTE}_{\textrm{nodual}}$ especially with few training samples.
\begin{figure}[t!]
	\setlength{\abovecaptionskip}{-0.0cm}
	\setlength{\belowcaptionskip}{-0.25cm}
	\centering 
	\includegraphics[width=0.68\linewidth]{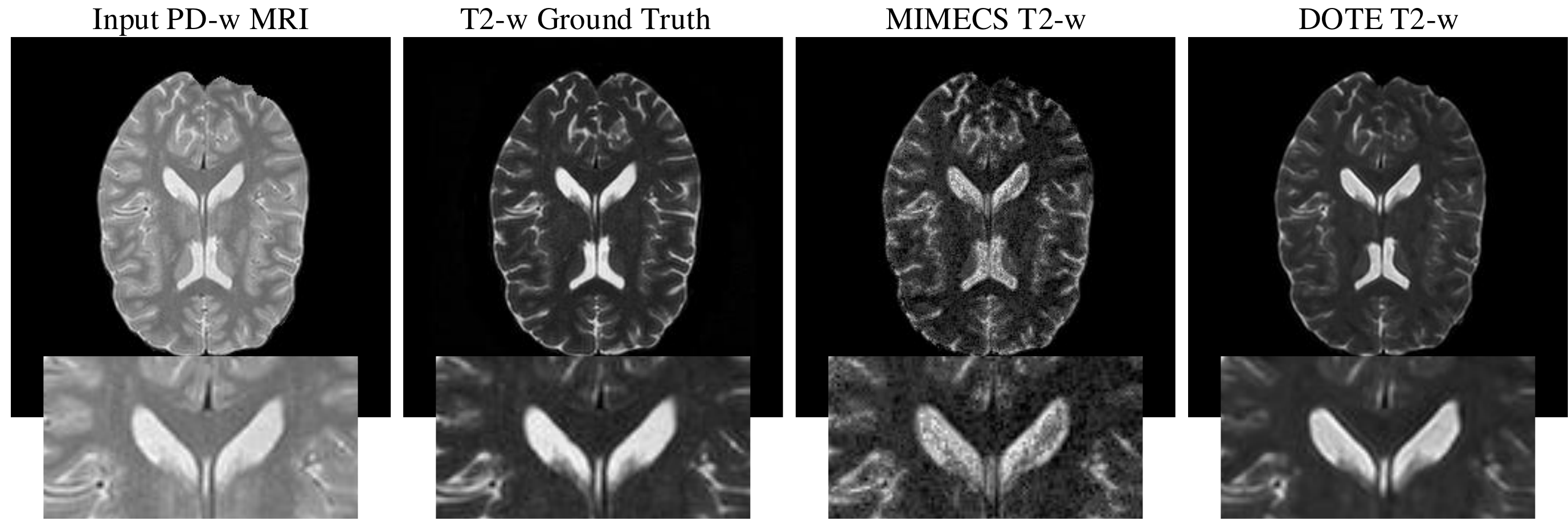} 
	\caption{Visual comparison of synthesized results using MIMECS and DOTE.}
	\label{fig4} 
\end{figure}
\begin{table}[t!]
	\scriptsize
	\setlength{\abovecaptionskip}{-0.0cm}
	\setlength{\belowcaptionskip}{-0.25cm}
	\centering 
	\begin{tabular}{|c|c|c|c|c|c|c|c|}
		\hline
		avg. & ScSR & Zeyde & NLSR & ANR & A+ & CSC-SR & DOTE\\
		\hline
		PSNR & 29.98 & 33.10 & 33.97 & 35.23 & 35.72 & 36.18 & \textbf{37.07}\\
		\hline
		SSIM & 0.9265 & 0.9502 & 0.9548 & 0.9568 & 0.9600 & 0.9651 & \textbf{0.9701}\\
		\hline
	\end{tabular}
	\caption{Quantitative evaluation: DOTE vs. other SR methods.} 
	\label{tab1}
\end{table}
\begin{table}[t!]
	\scriptsize
%	\fontsize{5.5}{6.3}\selectfont
	\setlength{\abovecaptionskip}{-0.0cm}
	\setlength{\belowcaptionskip}{-0.0cm}
	\centering 
	\begin{tabular}{|c|c|c|c|c|c|c|c|}
		\hline 
		avg. & $\textrm{DOTE}_{\textrm{nodual}}\frac{1}{4}$ & $\textrm{DOTE}_{\textrm{nodual}}\frac{1}{2}$ & $\textrm{DOTE}_{\textrm{nodual}}\frac{3}{4}$ & DOTE $\frac{1}{4}$ & DOTE $\frac{1}{2}$ & DOTE $\frac{3}{4}$\\
		\hline
		PSNR & 31.23 & 33.17 & 36.09 & \textbf{36.56} & \textbf{36.68} & \textbf{37.07} \\
		\hline
		SSIM & 0.9354 & 0.9523 & 0.9581 & \textbf{0.9687} & \textbf{0.9690} & \textbf{0.9701} \\
		\hline
	\end{tabular}
	\caption{Quantitative evaluation: DOTE vs. $\textrm{DOTE}_{\textrm{nodual}}$.} 
	\label{tab2}
\end{table}

\begin{figure}[t!]
	\setlength{\abovecaptionskip}{-0.0cm}
	\setlength{\belowcaptionskip}{-0.25cm}
	\centering 
	\includegraphics[width=1\linewidth]{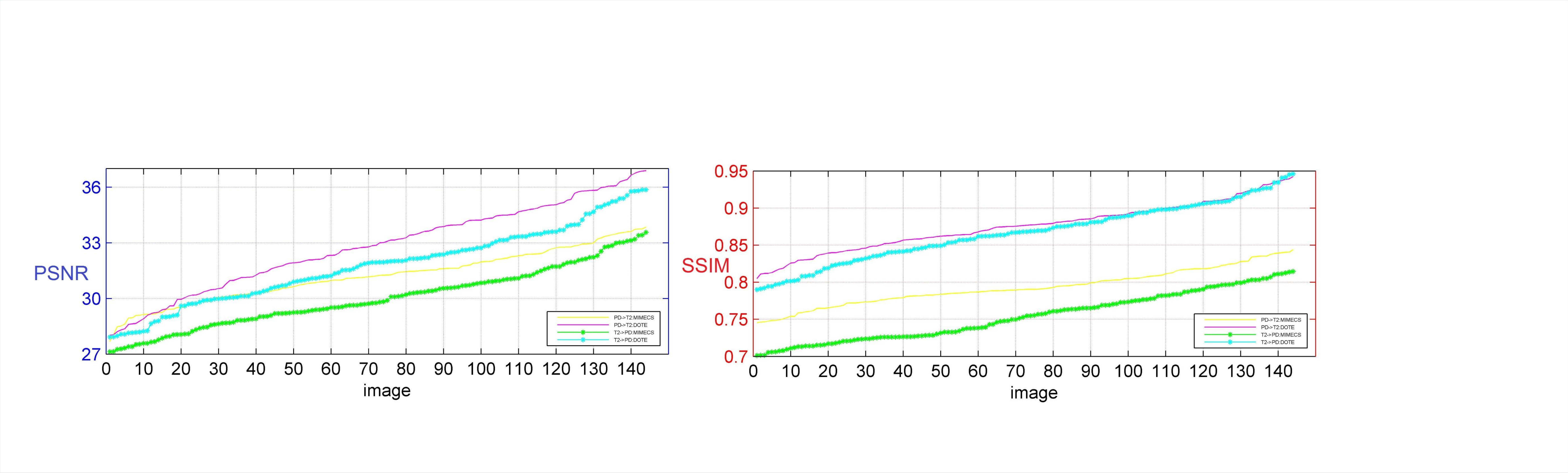} 
	\caption{CMS results: DOTE vs. MIMECS on the IXI dataset.}
	\label{fig5} 
\end{figure}

\textbf{Cross-Modality Synthesis.} For the problem of CMS, we evaluate DOTE and the relevant algorithms on both datasets involving six groups of experiments: (1) synthesizing T2-w image from PD-w acquisition and (2) \textit{vice versa}; (3) generating T1-w image from T2-w input, and (4) \textit{vice versa}. We conduct (1-2) experiments on the IXI dataset, while (3-4) are explored on the NAMIC dataset. The representative and state-of-the-art CMS methods, including Vemulapalli's method \cite{Vemulapalli} and MIMECS \cite{Roy} are employed to compare with our DOTE approach. We demonstrate visual and quantitative results in Fig. \ref{fig4}, Fig. \ref{fig5} and Table. {\ref{tab3}}, respectively. Our algorithm yields the best results against MIMECS and Vemulapalli for two datasets validating our claim of being able to synthesize better results through the expanded dual optimization.
\begin{table}[t!]
	\fontsize{6}{6.8}\selectfont
	\setlength{\abovecaptionskip}{-0.0cm}
	\setlength{\belowcaptionskip}{-0.0cm}
	\centering
	\begin{tabular}{|c||c|c|c|c|c|c|}
		\hline
		\multirow{3}{*}{Metric(avg.)} & \multicolumn{6}{c|}{NAMIC}\\
		\cline{2-7} & \multicolumn{3}{c|}{T1$->$T2} & \multicolumn{3}{c|}{T2$->$T1} \\
		\cline{2-7} & MIMECS & Vemulapalli & DOTE & MIMECS & Vemulapalli & DOTE \\
		\hline
		PSNR & 24.98 & 27.22 & \textbf{29.83} & 27.13 & 28.95 & \textbf{32.03}\\
		\hline
		SSIM & 0.8821 & 0.8981 & \textbf{0.9013} & 0.9198 & 0.9273 & \textbf{0.9301}\\
		\hline
	\end{tabular}
	\caption{CMS results: DOTE vs. other synthesis methods on the NAMIC dataset.}
	\label{tab3}
\end{table}

\section{Conclusion}
We presented a dual convolutional filter learning (DOTE) method which directly decomposes the whole image based on CSC, such that local neighbors are preserved consistently. The proposed dual mapping functions integrated with joint learning model form a closed loop that leverages the training data more efficiently and keeps a very stable mapping between image modalities. We applied DOTE to both image SR and CMS problems. Extensive results showed that our method outperforms other state-of-the-art approaches. Future work could concentrate on extending DOTE to higher-order imaging modalities like diffusion tensor MRI and to other modalities beyond MRI.


\begin{thebibliography}{4}
	\bibitem{Yang} Yang, J., Wright, J., Huang, T.S., Ma, Y.: Image super-resolution via sparse representation. IEEE TIP, 19(11), pp.2861-2873 (2010).
	
	\bibitem{Timofte} Timofte, R., De Smet, V., Van Gool, L.: A+: Adjusted anchored neighborhood regression for fast super-resolution. In: ACCV, pp. 111-126 (2014).	
	
	\bibitem{Vemulapalli} Vemulapalli, R., Van Nguyen, H., Kevin Zhou, S: Unsupervised cross-modal synthesis of subject-specific scans. In: IEEE ICCV, pp. 630-638 (2015).	
	
	\bibitem{Rousseau} Rousseau, F., Alzheimer’s Disease Neuroimaging Initiative: A non-local approach for image super-resolution using intermodality priors. MIA, 14(4), pp.594-605 (2010).
	
	\bibitem{Zeyde} Zeyde, R., Elad, M., Protter, M.,: On single image scale-up using sparse-representations. In: ICCS, pp. 711-730 (2010).
	
	\bibitem{Bristow} Bristow, H., Eriksson, A., Lucey, S.: Fast convolutional sparse coding. In: IEEE CVPR, pp. 391-398 (2013).
	
	\bibitem{He} He, D., Xia, Y., Qin, T., Wang, L., Yu, N., Liu, T., Ma, W.: Dual Learning for Machine Translation. NIPS, pp. 820-828 (2016).
	
	\bibitem{Wang} Wang, S., Zhang, L., Liang, Y., Pan, Q.: Semi-coupled dictionary learning with applications to image super-resolution and photo-sketch synthesis. In: IEEE CVPR, pp. 2216-2223 (2012).	
	
	\bibitem{Gu} Gu, S., Zuo, W., Xie, Q., Meng, D., Feng, X., Zhang, L.: Convolutional sparse coding for image super-resolution. In: IEEE ICCV pp. 1823-1831 (2015).	
	
	\bibitem{Timofte2} Timofte, R., De Smet, V., Van Gool, L.: Anchored neighborhood regression for fast example-based super-resolution. In: IEEE ICCV, pp. 1920-1927 (2013).	
	
	\bibitem{Roy} Roy, S., Carass, A., Prince, J.L.: Magnetic resonance image example-based contrast synthesis. IEEE TMI, 32(12), pp.2348-2363 (2013).		
\end{thebibliography}
\end{document}